\documentclass{zapiski}
\makeatletter

\usepackage[T2A]{fontenc}
\usepackage[russian, english]{babel}

\usepackage[normalem]{ulem} 
\usepackage{xcolor}
\usepackage{etoolbox} 
\usepackage{times}
\usepackage{url}
\usepackage{latexsym}
\usepackage{multirow}
\usepackage{lipsum}
\usepackage{graphicx}
\usepackage{amssymb}
\usepackage{hyperref}
\usepackage{multirow}
\usepackage{booktabs}
\usepackage{tabularx}
\usepackage{caption}
\usepackage{cite}
\usepackage{listings}
\usepackage{svg}

\usepackage{paralist}
\usepackage{subcaption}
\usepackage{xcolor,colortbl}
\usepackage{array}
\usepackage{setspace}

\usepackage{placeins}

\usepackage{tikz}
\usepackage{pgfplots}

\newcolumntype{H}{>{\setbox0=\hbox\bgroup}c<{\egroup}@{}}
\newcolumntype{Z}{>{\setbox0=\hbox\bgroup}c<{\egroup}@{\hspace*{-\tabcolsep}}}

\usepackage{float}
\restylefloat{table}

\english

\begin{document}

\title[When LLM uncertainty is justified]{When an LLM is apprehensive about its answers - and when its uncertainty is justified}


\author[Sychev]{Petr Sychev}
\address[P. Sychev]{Skolkovo Institute of Science and Technology (Skoltech) / Moscow, Russia\\National Research University Higher School of Economics / Moscow, Russia}
\email{petr.sychev@skoltech.ru}

\author[Goncharov]{Andrey Goncharov}
\address[A. Goncharov]{Skolkovo Institute of Science and Technology (Skoltech) / Moscow, Russia}
\email{Andrey.Goncharov@skoltech.ru}

\author[Vyazhev]{Daniil Vyazhev}
\address[D. Vyazhev]{Skolkovo Institute of Science and Technology (Skoltech) / Moscow, Russia\\National Research University Higher School of Economics / Moscow, Russia}
\email{daniel.vyazhev@skoltech.ru}

\author[Khalafyan]{Edvard Khalafyan}
\address[E. Khalafyan]{Moscow Institute of Physics and Technology / Moscow, Russia\\Skolkovo Institute of Science and Technology (Skoltech) / Moscow, Russia}
\email{khalafyan.ea@phystech.edu}

\author[Zaytsev]{Alexey Zaytsev}
\address[A. Zaytsev]{Skolkovo Institute of Science and Technology (Skoltech) / Moscow, Russia}
\email{a.zaytsev@skoltech.ru}

\begin{abstract}
Uncertainty estimation is crucial for evaluating Large Language Models (LLMs), particularly in high-stakes domains where incorrect answers result in significant consequences.
Numerous approaches consider this problem, while focusing on a specific type of uncertainty, ignoring others.
We investigate what estimates, specifically token-wise entropy and model-as-judge (MASJ), would work for multiple-choice question-answering tasks for different question topics.
Our experiments consider three LLMs: Phi-4, Mistral, and Qwen of different sizes from 1.5B to 72B and $14$ topics.
While MASJ performs similarly to a random error predictor, the response entropy predicts model error in knowledge-dependent domains and serves as an effective indicator of question difficulty: for biology ROC AUC is $0.73$. This correlation vanishes for the reasoning-dependent domain: for math questions ROC-AUC is $0.55$. 
More principally, we found out that the entropy measure required a reasoning amount. Thus, data-uncertainty related entropy should be integrated within uncertainty estimates frameworks, while MASJ requires refinement. 
Moreover, existing MMLU-Pro samples are biased, and should balance required amount of reasoning for different subdomains to provide a more fair assessment of LLMs performance.
\end{abstract}


\maketitle

\section{Introduction}

LLMs have succeeded in various tasks, demonstrating impressive capabilities in generating human-like text and solving complex problems. However, their ability to accurately assess uncertainty in their predictions remains a significant challenge~\cite{kadavath2022language}. 
This gap between a model's confidence and its actual correctness, often called the confidence gap \cite{chhikara2025mindconfidencegapoverconfidence}, poses critical risks, particularly in high-stakes applications where overconfidence in incorrect answers leads to severe consequences: in domains such as health, law~\cite{Dahl_2024}, education, or economics an LLM's overestimation of its accuracy could result in harmful decisions or misinformation. 
Understanding and decreasing this confidence gap is therefore essential for ensuring the safe and reliable deployment of LLMs. 

Addressing this limitation has led to the development of numerous uncertainty estimation techniques. 
Most of them belong to one of two problem statements.
They either focus on binary classification or rely on rich signals from long, free-text responses~\cite{yang2024logulongformgenerationuncertainty}.
These methods often generalize poorly for concise multiple-choice questions.
\begin{figure}
    \centering
    \includegraphics[width=1\linewidth]{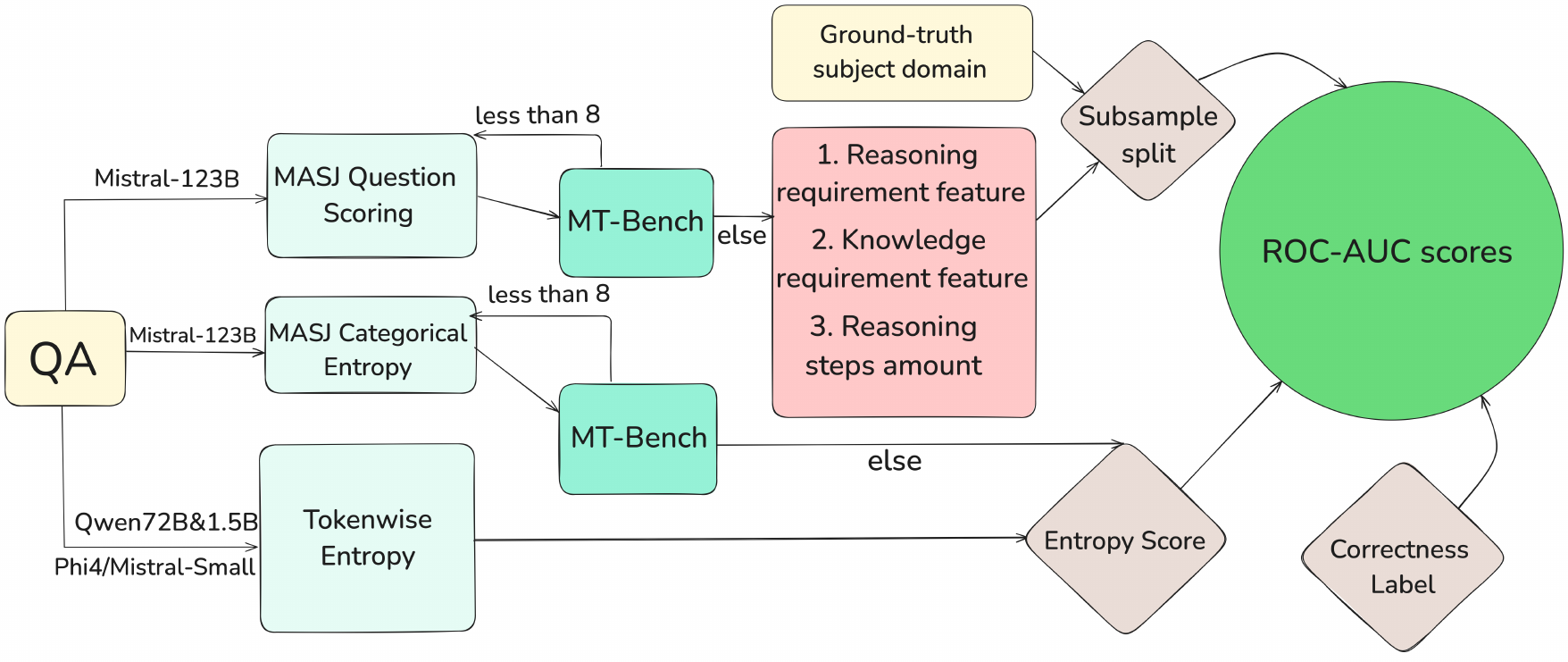}
    \caption{Question complexity evaluation pipeline. See more details in the methods section}
    \label{fig:difficult_questions_scheme}
\end{figure}

Moreover, existing methods tend to answer how good an overall confidence estimate is, ignoring why uncertainty appeared in a specific case. 
While classic decomposition of uncertainty to aleatoric (data) and epistemic (model)~\cite{hullermeier2021aleatoric} also holds for LLMs, the problem is challenging~\cite{huang2025survey}.
For example, the authors in~\cite{abbasi2025believe} consider only epistemic uncertainty.

To address these gaps, we propose a pipeline that enables detailed investigation of domain-specific or complexity-labeled datasets with multiple-choice answers. 
This framework allows us to explore how commonly used uncertainty estimation methods, particularly entropy-based approaches, perform under varying question topics and levels of required reasoning. 
We can also leverage an auxiliary LLM to generate labels for reasoning, knowledge requirements, and the number of reasoning steps. 
The introduction of the pipeline led to a more detailed investigation of datasets commonly used for benchmarking LLMs and their uncertainty estimates.
In more detail, our main contributions are the following:
\begin{itemize}
    \item We present a fully automated pipeline for evaluating uncertainty estimation approaches within the MMLU-Pro setting. It is presented in Figure~\ref{fig:difficult_questions_scheme} and allows datasets or tested methods to be varied. The associated work is available on Github\footnote{\url{https://github.com/LabARSS/question-complextiy-estimation}}.
    \item We propose using data uncertainty estimates based on token-wise entropy and model uncertainty estimates based on MASJ. Thus, our approach covers two main components of uncertainty estimation: data and model one, given that MMLU-Pro answers are typically short and provide limited information on model confidence.
    \item Within the developed pipeline, we consider different subsets of the MMLU-Pro dataset with splits by domain and reasoning requirement estimated via MASJ proposed in this work.
    \item Our experimental evidence that includes consideration of four models (Phi-4, Mistral-Small-24B-Instruct, Qwen-1.5 B, Qwen-72B) suggest that the entropy predicts well errors of an LLM if the amount of required reasoning is small. Moreover, its quality increases if the model size increases. Unlike MASJ provides much weaker results, being unable to identify error patterns. Further progress here would be possible with more reasoning steps during an uncertainty estimate with an entropy-based approach on top of it.
    \item We also observe that the existing dataset MMLU-Pro has internal biases with the complexity of the question, the same as in~\cite{gema2025mmlu}, and the required reasoning amount for them varies significantly depending on the topic. Thus, more advanced and less biased datasets are required for a fair measurement of the progress of LLMs. Additional human annotation of questions in a dataset could identify further gaps in LLMs' capabilities.
\end{itemize}

\section{Related Works}

Tests have been a popular way to estimate an individual's proficiency for more than 500 years~\cite{buyske2005optimal}. 
After their emergence, LLMs have also been tested extensively and compared to each other or humans~\cite{sizov2024analysing}. 
Several papers established benchmarks, with natural ones being QA datasets: an LLM receives a prompt and should generate a response~cite{rogers2023qa}.
Different questions exist, but automated evaluation of answers is available only for multiple-choice questions.
For LLMs, slight changes in the way the questions are asked can significantly affect the accuracy of a model, probably due to data leakage and format overfitting~\cite{elhady2025wicked},
other options are much more expensive and challenging.
One of the recent options that takes into account possible leakages - and, on the other hand, is widely used is an expanded version of the MMLU dataset, called MMLU-Pro~\cite{wang2025mmlu} with about $0.56$ accuracy for Llama-70B model.
The presented $\sim 12 000$ questions span $13$ diverse topics, including STEM and other areas of scientific studies.
The complexity, volume, and diversity of the dataset provide a strong foundation for our study of how different aspects of the questions affect the probability of correctly answering them.

In related works, datasets such as ARC-AI2~\cite{clark2018think} and GPQA~\cite{rein2023gpqagraduatelevelgoogleproofqa} have been widely used for evaluating multiple-choice QA in the scientific domain as well. However, ARC-AI2 is often criticized for its relatively low difficulty, while GPQA, though highly challenging, suffers from a dataset size of fewer than $500$ samples, which restricts its utility for robust evaluation. To address these limitations and ensure a comprehensive assessment, we focus our final experiments on MMLU-Pro, which offers a larger scale and a more balanced difficulty level for evaluating model performance in scientific reasoning tasks.

MMLU-Pro is suited for evaluating a model's ability to identify uncertainty in the generated output and signal potential errors, which addresses a fundamental question in AI safety. 
In a broader context, we can refer to this effect as hallucination~\cite{huang2025survey}; in a more classic context, it is a prediction error~\cite{fadeeva2023lm}.
Another related approach is assessing the question difficulty, provided in~\cite{gabburo2024measuring}, but suitable there only for an RAG scenario. Universal Model-as-judge (MASJ) approaches use available models or a bigger and more powerful model to evaluate LM outputs~\cite{zheng2023judging} and can be applied in a broader context.

What makes the problem more difficult is the diverse nature of possible uncertainty sources.
Classic works decompose uncertainty into the model part, related to lack of knowledge of a model about a specific output, and the data part, related to the complexity of the data itself~\cite{hullermeier2021aleatoric}.
For natural language, model uncertainty refers to the familiarity of a specific domain and the presence of relevant content in a training sample.
Data uncertainty refers to noise in the example or the overall complexity of a question at hand.
While broader classification exists with five different features for a generated question~\cite{wan2024sciqag}, we follow this approach while identifying the considered parts of the uncertainty. Using entropy-based measures, such as semantic entropy~\cite{kuhn2023semanticuncertaintylinguisticinvariances} and word-sequence entropy~\cite{wang2024wordsequenceentropyuncertaintyestimation}, is highly effective for evaluating uncertainty in LLMs because these methods capture the variability and confidence of model predictions in a structured way. Semantic entropy, as proposed, quantifies uncertainty by considering the meaning of generated text, making it robust to paraphrasing and linguistic variations. Similarly, word-sequence entropy demonstrates that word-sequence entropy effectively measures uncertainty in open-ended tasks such as medical QA, providing insights into model confidence and reliability. In our case, we use the token-wise entropy approach~\cite{kadavath2022language} due to its simplicity and suitability for MMLU cases with multiple-choice short answers.
For model uncertainty, we consider a variation of the MASJ approach with a specific prompt~\cite{zheng2023judging}, ~\cite{chern2024largelanguagemodelstrusted}.

This~\cite{shorinwa2024surveyuncertaintyquantificationlarge} survey paper provides a comprehensive taxonomy of uncertainty estimation methods in LLMs, highlighting state-of-the-art approaches such as evaluation via LLM and tokenwise entropy.
For an MMLU-type dataset, we would be able to identify how different types of uncertainties are combined for typical questions in different subjects and how our measurement corresponds to the estimate of the required amount of reasoning for specific questions.
These results would help identify hallucinations and errors in language model outputs and provide insights into how to reduce the frequency of these effects by training the model using additional data and varying the model size.

\section{Methods}

\subsection{General Pipeline}

\textbf{Input.} We consider a set of questions $Q = \{q_i \}_{i = 1}^n$ and a set of answers $A = \{a_i \}_{i = 1}^n$ of size $n$ with generated answers $G = \{g_i \}_{i = 1}^n$.
The questions are augmented with true topic labels $T = \{t_i \}_{i = 1}^n$ and correctness of the answers $Y = \{y_i \}_{i = 1}^n$, where $y_i$ is binary value which means equality of ground-truth answer $a_i$ and generated answer $g_i$

\textit{Uncertainty estimation.} Our procedures estimate the uncertainty of a model response given a question and a corresponding answer $U = \{u_i \}_{i = 1}^n$, each $u_i = f(q_i, g_i)$ and depends on the LLM used.
We additionally prompt models to obtain an estimation of the reasoning requirements $R = \{r_i \}_{i = 1}^n$. 

\textit{Uncertainty validation.} Finally, given all this information, we compare the uncertainty estimates $U$ with the correctness labels $Y$ to obtain the ROC-AUC values and other quality measures. Additionally, we examine how good specific uncertainty estimates $U$ are for separate topics $T$ or the required amount of reasoning $R$.

\textit{Technical details.}
For each question in the dataset, we prompt a considered model for an answer and record the entropy for the answer. We discard answers with incorrect formatting that appear in less than $5\%$ of the questions.

In model-as-judge (MASJ), we use a large 123B Instruct Mistral model \texttt{Mistral-Large-Instruct-2411} to obtain estimated complexity according to the level of education.
The same MASJ technique with the same large model was utilized to obtain the reasoning estimate. We ask a binary question of the deep reasoning required and how many steps it would take: low, medium, high.
Specific prompts are available in the Appendix~\ref{prompts}.

For a more detailed analysis, we split the whole set of questions to topics using available labels or reasoning and educational level complexity obtained via MASJ.

\subsection{Uncertainty Estimations}

We consider two reliable approaches for uncertainty estimation: the entropy-based procedure~\cite{kadavath2022language} and the model-as-a-judge~\cite{zheng2023judging}.
Details on both of them are provided below.

\subsubsection{Entropy Uncertainty Estimation}

Entropy-based uncertainty estimation is a fundamental approach for evaluating LLMs confidence in their predictions~\cite{fadeeva2024fact}. 
In MMLU, the expected answer is a number with one or two tokens length.
So, we use token-wise entropy from the model's output logits.

Our implementation computes entropy via a two-step procedure.
Firstly, we obtain the logits $\mathbf{z} = (z_1, \ldots, z_k)$ from the LLM's final layer for each token position, where $k$ equals the vocabulary size. 
Secondly, we convert these logits into a probability distribution using the softmax function:
$$
p_i = \frac{e^{z_i}}{\sum_{j = 1}^k e^{z_j}}.
$$
Next, we calculate the entropy of this distribution:
$$
u = -\sum_{i  = 1}^k p_i \log p_i.
$$

This approach aligns with the work~\cite{kadavath2022language}, which showed that token-level entropy correlates well with model uncertainty and performance. 
When a model is confident in its next-token prediction, the probability mass concentrates on a small subset of tokens, resulting in a low entropy. Conversely, when uncertain, the model distributes probabilities more evenly across the vocabulary, producing higher entropy values.

We hypothesize that high-entropy responses generally correspond to questions where the model lacks the necessary knowledge or reasoning capacity. By analyzing the relationship between entropy and model performance across domains and difficult reasoning complexity, we can identify specific conditions under which LLMs are most prone to overconfidence or appropriate uncertainty calibration.

\subsubsection{MASJ Uncertainty Estimation}

The model-as-judge (MASJ) paradigm represents an alternative approach to uncertainty estimation that leverages an LLM's ability to evaluate a given response. 
Unlike entropy-based methods that rely on a probability distribution across output tokens, MASJ employs prompt-based techniques.

Our implementation of MASJ follows a methodology similar to the approach described in the MT-Bench evaluation framework~\cite{zheng2023judging}. This approach extracts explicit confidence assessments that may capture uncertainty beyond what is reflected in the raw probability distribution: including broader reasoning about the model's knowledge boundaries, the ambiguity of questions, and the model's awareness of its own limitations. In instances where MT-Bench assigned a score below 8 on a scale of 1 to 10, we decided to regenerate the estimate.

\section{Results}

\subsection{Experimental setup}

\subsubsection{\emph{Datasets}}

\paragraph{MMLU-Pro}
In this study, we utilize the MMLU-Pro dataset~\cite{wang2025mmlu}, a more challenging and robust variant of the well-known Massive Multitask Language Understanding (MMLU) benchmark. The original MMLU dataset was introduced by Hendrycks et al.~\cite{hendrycks2020measuring} and has been widely used to evaluate language model performance across diverse domains. MMLU-Pro builds upon this benchmark by incorporating a filtered and refined subset of the original questions, also introducing newly generated question-answer pairs to enhance difficulty and robustness.

The dataset consists of approximately $12000$ multiple-choice questions, with $10$ answer choices per question in the majority of cases (approximately $ 10000$ instances), while the remaining $ 2000$ instances contain between $3$ to $9$ answer options. 
The dataset spans 14 distinct scientific categories, covering a broad range of disciplines with the number of topics by discipline provided in Table~\ref{tab:mmlu_pro_distr}.
Examples of questions from different categories are available in Appendix~\ref{sec:question_examples}.

\begin{figure}
    \footnotesize\setlength{\tabcolsep}{2.8pt}
    \begin{tabular}{|c|c|c|}
    \hline
        Category & Sample Count \\
        \hline
        Mathematics & 1351 \\
        Physics & 1299 \\
        Chemistry & 1132 \\
        Law & 1101 \\
        Engineering & 969 \\
        Economics & 844 \\
        Health & 818 \\
        Psychology & 798 \\
        Business & 789 \\
        Biology & 717 \\
        Philosophy & 499 \\
        Computer Science & 410 \\
        History & 381 \\
        Other & 924 \\
        \hline
    \end{tabular}
    \captionof{table}{MMLU-Pro Category Distribution}
    \label{tab:mmlu_pro_distr}
\end{figure}

The motivation for using MMLU-Pro for this problem is its increased difficulty and improved robustness compared to the original MMLU dataset. 
Official reports for Phi-4 and Mistral-Small-24B also used the extended dataset to mitigate data leakage.
Thus, MMLU-Pro is a reliable and appropriate benchmark for evaluating our models' generalization and reasoning capabilities.

\subsubsection{\emph{Models.}} For QA and evaluation of the entropy, we use decoder-only LLMs Mistral (\textit{Mistral-Small-24B-Base-2501})~\cite{jiang2023mistral}, Phi-4~\cite{abdin2024phi} and Qwen models~\cite{qwen2025qwen25technicalreport}. 

\subsection{Reasoning and knowledge-based complexity}

We hypothesize that different questions validate different abilities of a model. Some focus more on the requirement to know specific things, while others consider a model's ability to reason within a specific topic.  

\begin{figure}[t!]
    \centering
    \begin{subfigure}[t]{0.5\textwidth}
        \centering
        \includegraphics[width=\linewidth]{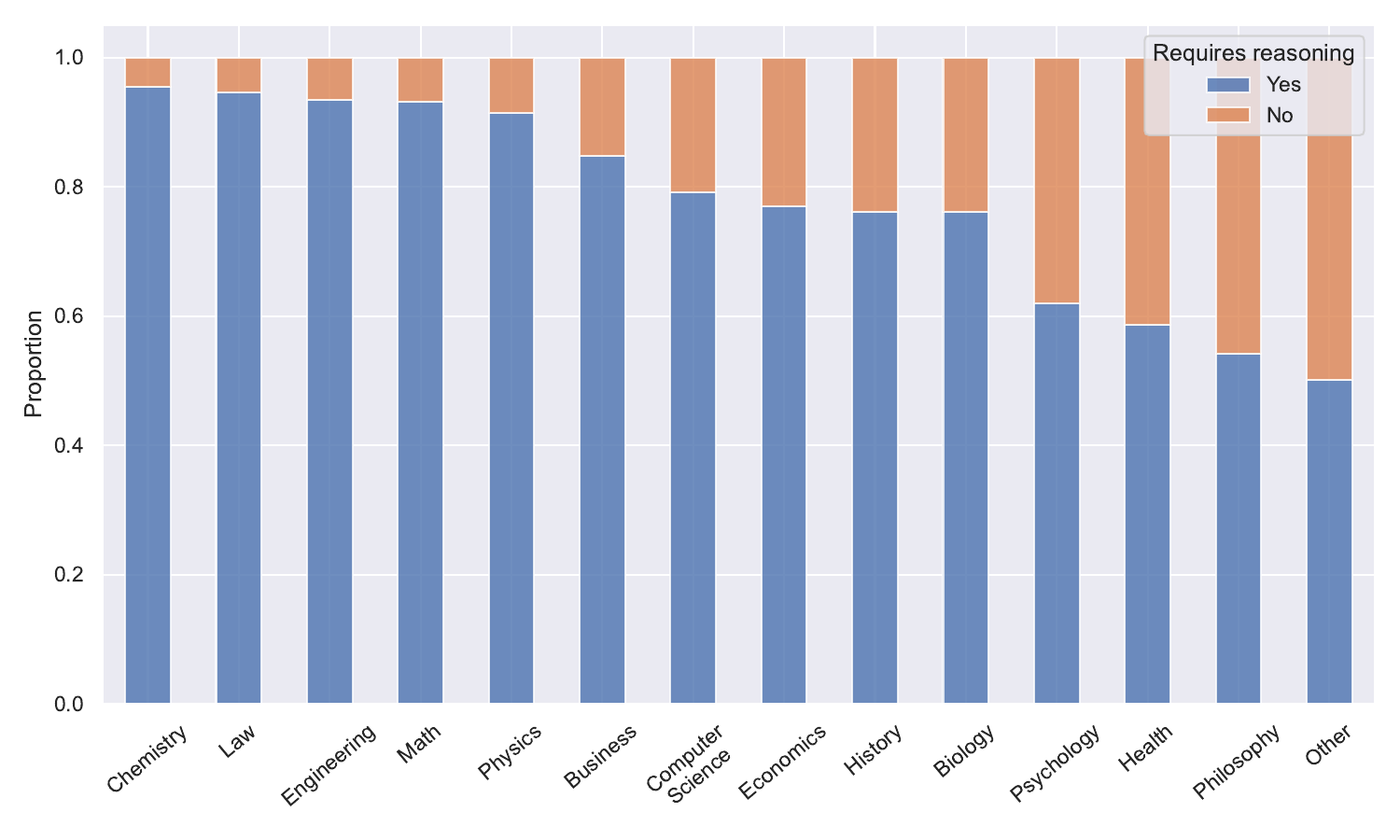}
        \caption{Distribution of questions that require complex reasoning}
    \end{subfigure}%
    ~ 
    \begin{subfigure}[t]{0.5\textwidth}
        \centering
        \includegraphics[width=\linewidth]{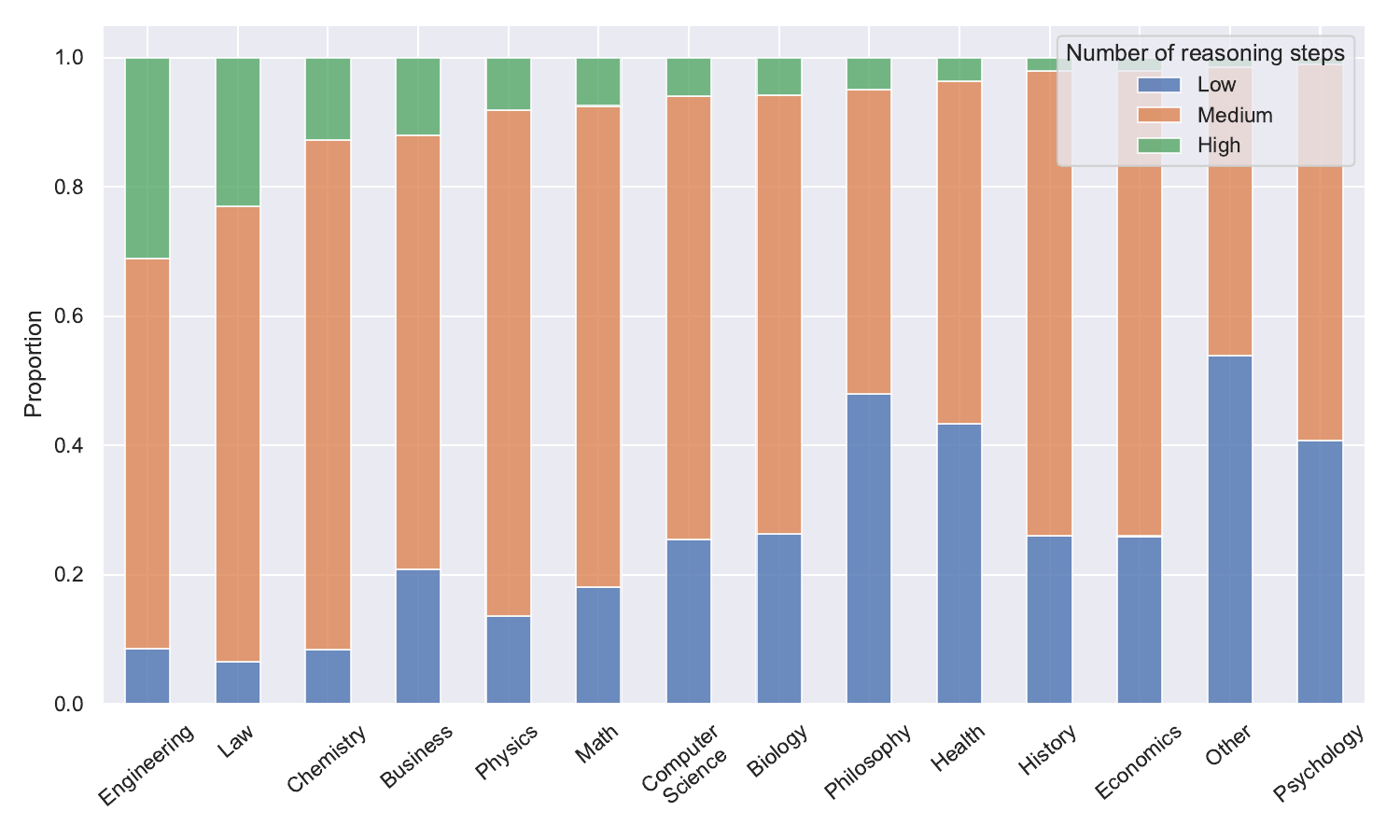}
        \caption{Distribution of the required number of reasoning steps}
    \end{subfigure}
    \caption{Estimation by MASJ of the required reasoning amount. Better to view in zoom}
    \label{fig:reasoning_distribution}
\end{figure}

To obtain these properties for specific questions, we obtained estimates via the model-as-judge.
Figures \ref{fig:reasoning_distribution} present the question type by the requirement to reason and the estimate for the required number of reasoning steps.

The distribution of questions over topics is diverse, with the reasoning requirement attaining the maximum share of about $0.9$ for engineering and only $0.5$ for philosophy. The number of required steps also varies: again, for engineering, the model estimates the number of questions with a high number of required steps, but for psychology and philosophy, this number is almost zero.

\subsection{Distribution of uncertainty measures}

\begin{figure}[t!]
    \centering
    \begin{subfigure}[t]{0.5\textwidth}
        \centering
        \includegraphics[width=\linewidth]{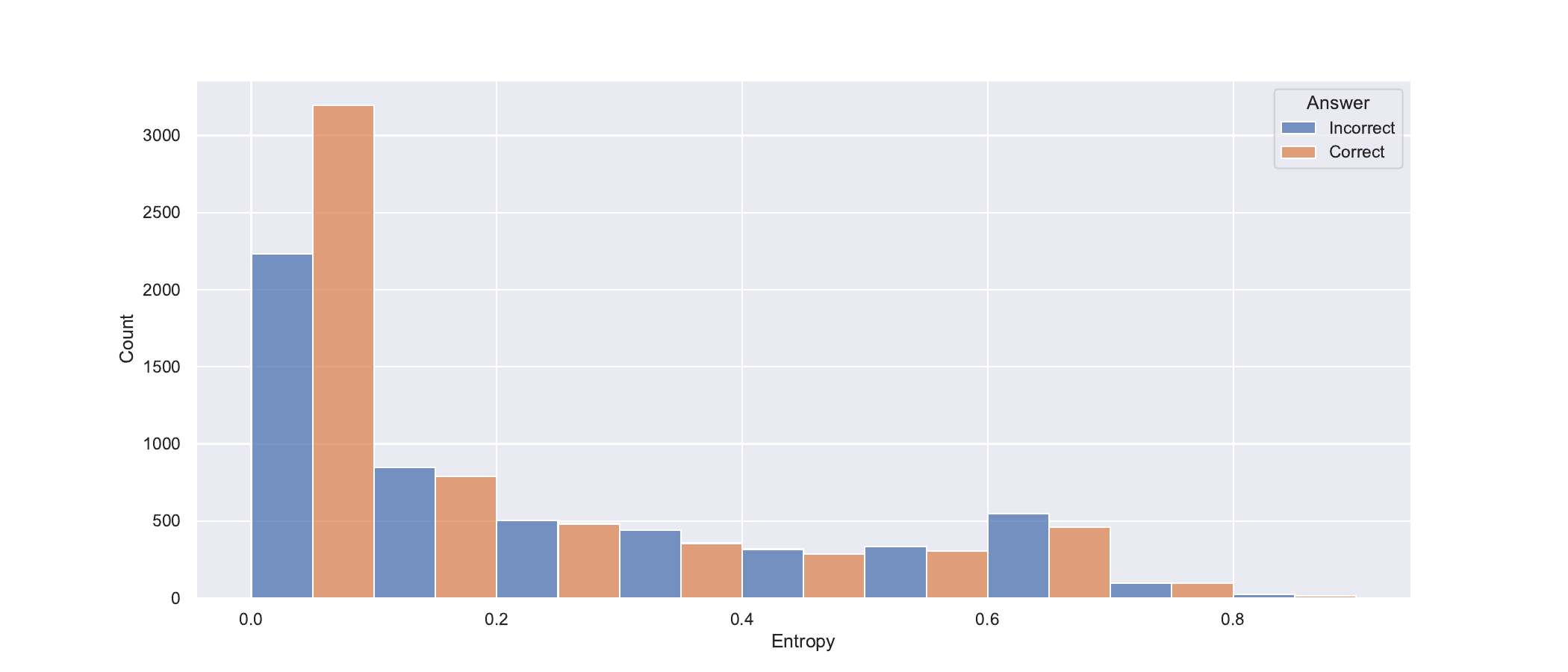}
        \caption{Phi-4 model} 
    \end{subfigure}%
    ~ 
    \begin{subfigure}[t]{0.5\textwidth}
        \centering
        \includegraphics[width=\linewidth]{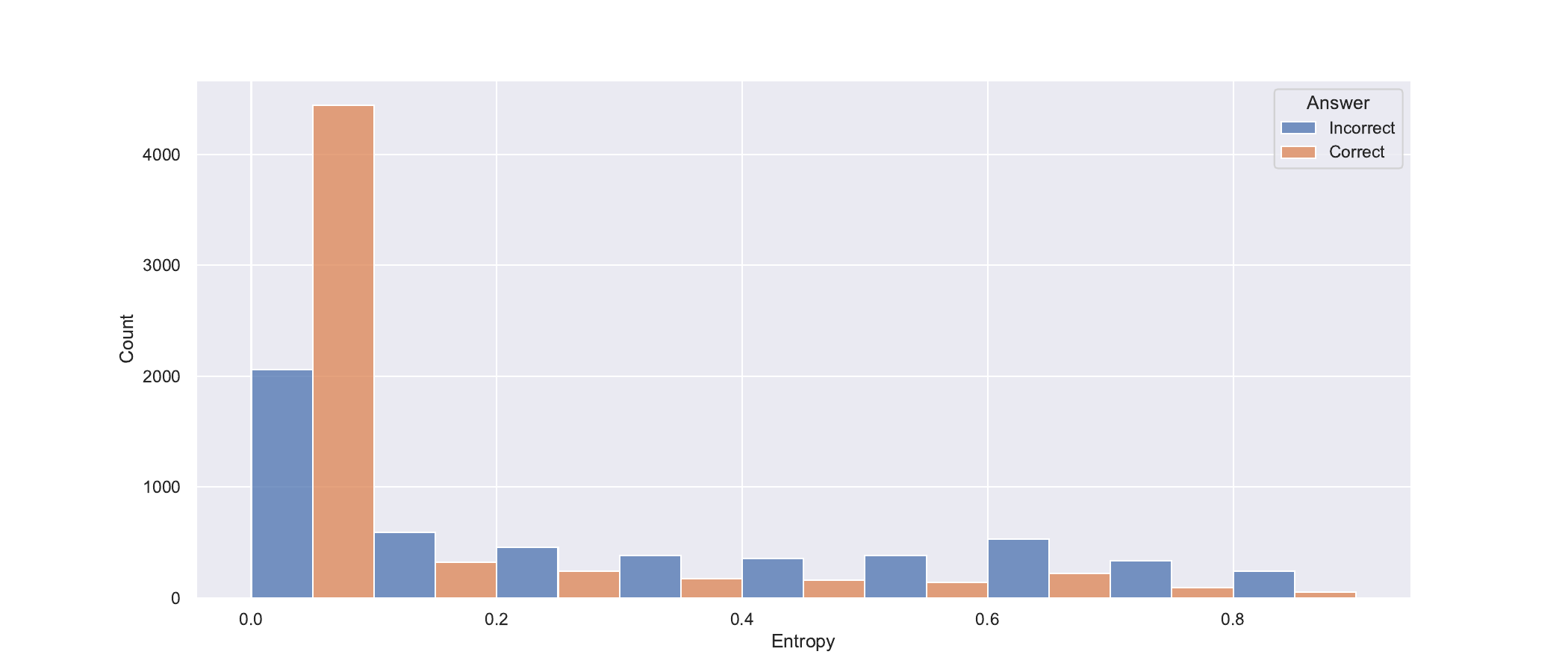}
        \caption{Qwen-72B model} 
    \end{subfigure}
    
    \vspace{1em}
    
    \centering
    \begin{subfigure}[t]{0.5\textwidth}
        \centering
        \includegraphics[width=\linewidth]{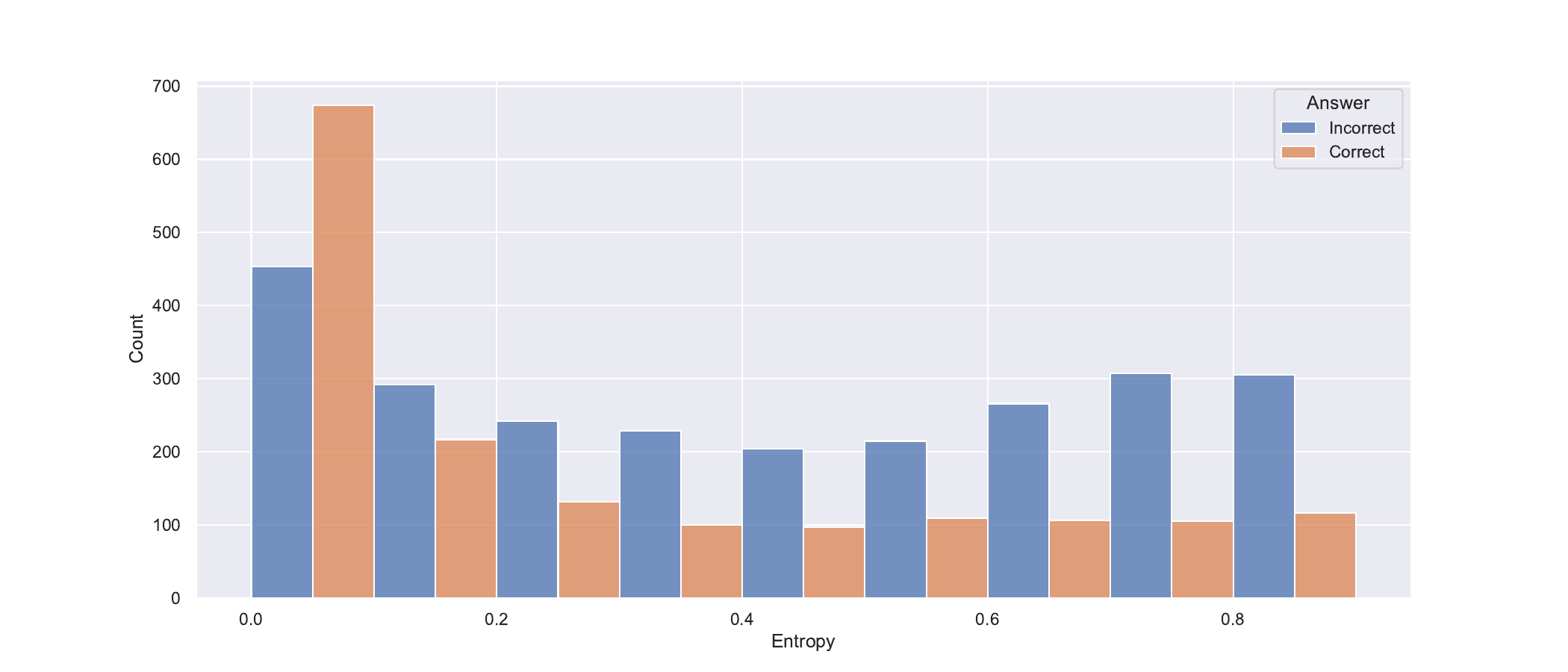}
        \caption{Qwen-1.5B model} 
    \end{subfigure}
    
    \caption{Entropy distribution of answers. Best viewed when zoomed in}
    \label{fig:entropy_distribution}
\end{figure}

For Phi-4 and Qwen we present the distribution of entropy values in Figures~\ref{fig:entropy_distribution}.
The distribution of entropy is presented for correct and incorrect answers by a model. 
We can see near-zero entropy values are much more often observed for correct answers for Phi-4 and Qwen. 
Thus, entropy only partially explains why the answers from a model are wrong, and the quality of this score depends on the model used.

Both numerical and nominal Model-as-judge scores also relate little to the complexity, as the corresponding measured ROC AUC values are $0.49$ for both models, indicating nearly random predictions.
Thus, MASJ uncertainty estimates focus on different aspects of an LLM's confidence, which is irrelevant to question complexity.

\begin{figure}
    \centering
    \includegraphics[width=1\linewidth]{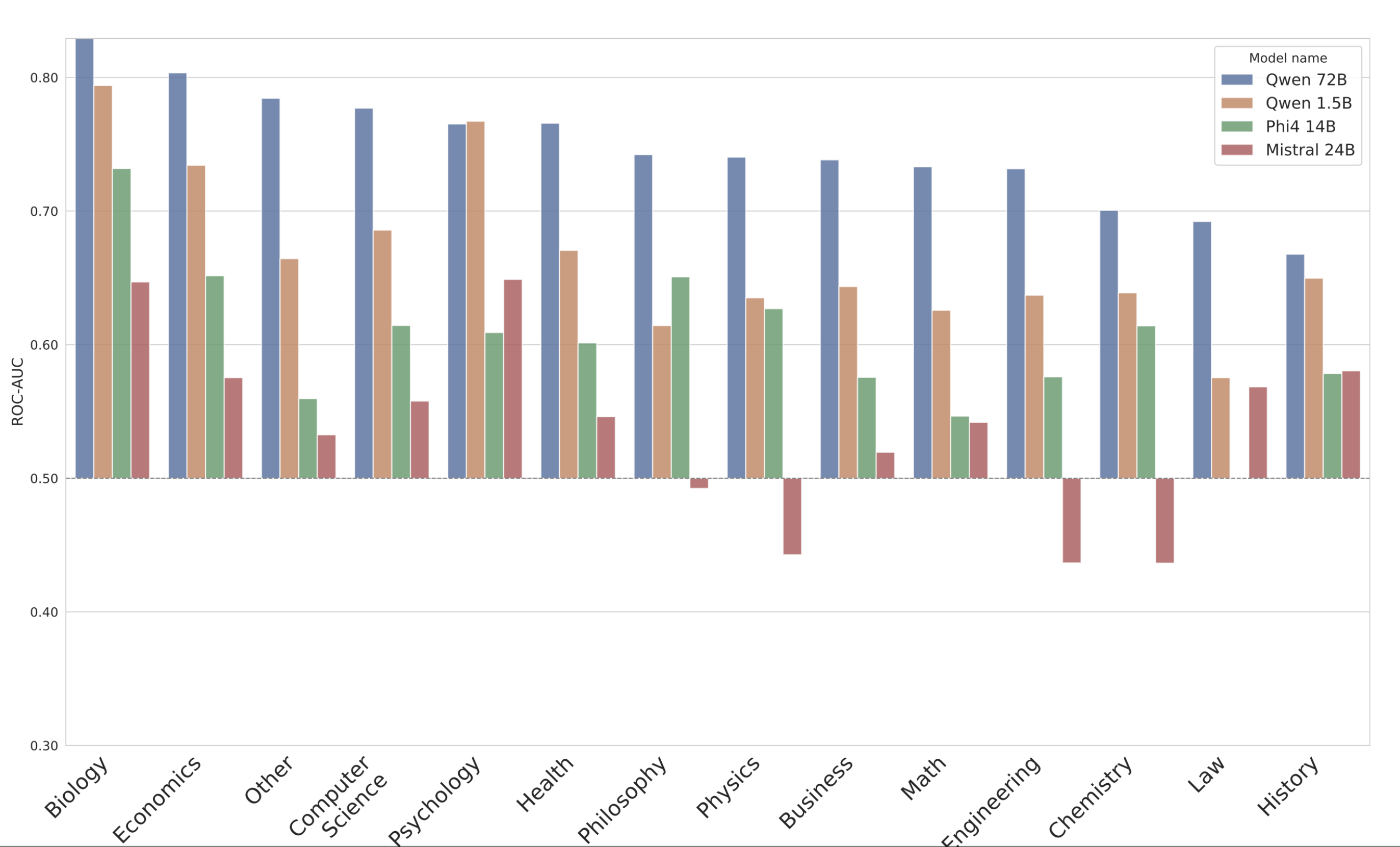}
    \caption{ROC-AUC for error prediction by subject for four different LLMs}
    \label{fig:models_compare_roc_auc}
\end{figure}

\begin{table}[htbp]
\centering
\renewcommand{\arraystretch}{1.2}
\begin{tabular}{lcccc}
\hline
Category & Phi-4 & Mistral & Qwen 1.5B & Qwen 72B \\
\hline
Biology & \textbf{0.73} & \textbf{0.65} & \textbf{0.79} & \textbf{0.83} \\
Economics & 0.65 & 0.58 & 0.73 & 0.80 \\
Philosophy & 0.65 & \textit{0.49} & 0.61 & 0.74 \\
Physics & 0.63 & \textbf{0.44} & 0.64 & 0.74 \\
Computer Science & 0.61 & 0.56 & 0.69 & {0.78} \\
Psychology & 0.61 &\textbf{0.65} & 0.77 & 0.77 \\
Chemistry & 0.61 & 0.44 & 0.64 & 0.70 \\
Health & 0.60 & 0.55 & 0.67 & 0.77 \\
Business & 0.58 & 0.52 & 0.64 & 0.74 \\
History & 0.58 & 0.58 & 0.65 & \textit{0.67} \\
Engineering & 0.58 & 0.44 & 0.64 & 0.73 \\
Other & 0.56 & 0.53 & 0.66 & {0.78} \\
Math & 0.55 & 0.54 & 0.63 & 0.73 \\
Law & \textit{0.50} & 0.57 & \textit{0.58} & 0.69 \\
\hline
No Reasoning & \textbf{0.59} & \textbf{0.59} & \textbf{0.72} & \textbf{0.79} \\
Yes Reasoning & 0.56 & \textit{0.52} & 0.68 & 0.76 \\
Reasoning (Low) & \textbf{0.59} & 0.58 & 0.71 & \textbf{0.79} \\
Reasoning (Med) & 0.57 & \textit{0.52} & 0.69 & 0.76 \\
Reasoning (High) & \textit{0.53} & 0.53 & \textit{0.61} & \textit{0.73} \\
\hline
Overall & 0.58 & 0.52 & 0.70 & 0.77 \\
\hline
\end{tabular}
\caption{ROC AUC performance comparison across categories and models}
\label{tab:rocauc}
\end{table}

\subsection{Complexity prediction via uncertainty estimate by category and question type}

\paragraph{Complexity prediction by category.} Figure \ref{fig:models_compare_roc_auc} show the ROC-AUC values for entropy for all four models. Values above or below $0.5$ indicate that the prediction is better than random. For most topics, the values are well above this threshold. In psychology and biology, we observe ROC-AUC values of $0.77/0.83$ (Qwen 72B) compared to $0.61/0.73$ (Phi-4) and $0.65/0.65$ (Mistral), with these improvements correlating across models.

\begin{figure}
    \centering
    \includegraphics[width=1\linewidth]{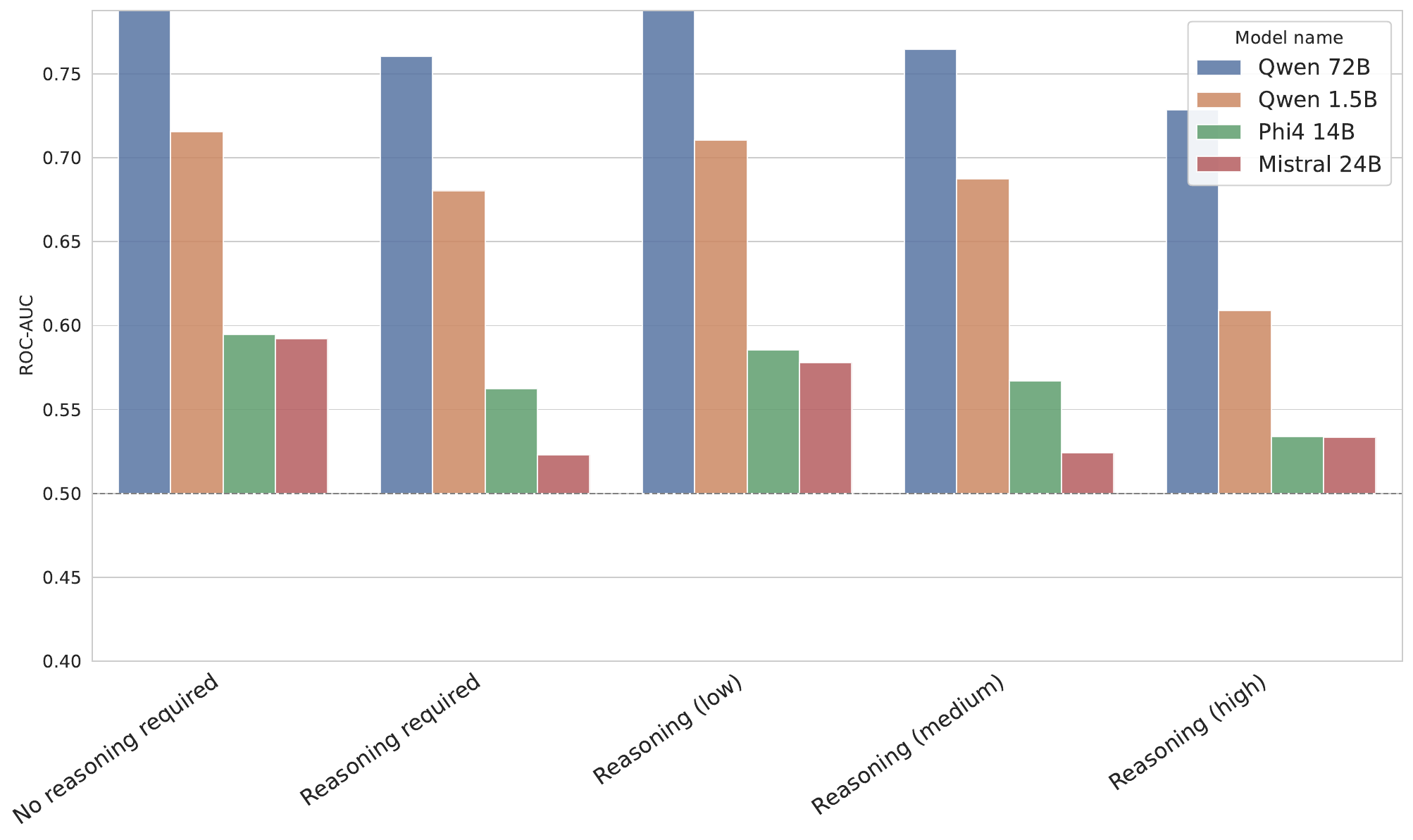}
    \caption{ROC-AUC by reasoning for four different LLMs}
    \label{fig:models_compare_reasoning}
\end{figure}

\paragraph{Complexity prediction by reasoning requirement.} 
Figure \ref{fig:models_compare_reasoning} present similar results, but for the split of the questions by reasoning complexity. Table \ref{tab:rocauc} shows aggregated ROC AUC scores across categories and reasoning levels. 

We see meaningful results of using the reasoning estimates for all questions without categorization by subject. Low entropy is a marginally better predictor of the truthfulness of the answers to questions that did not require complex reasoning and vice versa for models. We also observe the estimate of the number of reasoning steps as a less reliable metric, with Mistral favoring a higher number of reasoning steps. 
However, we can attribute it to the lower precision of such estimation provided by the MASJ approach. 

\begin{figure}
    \centering
    \includegraphics[width=1\linewidth]{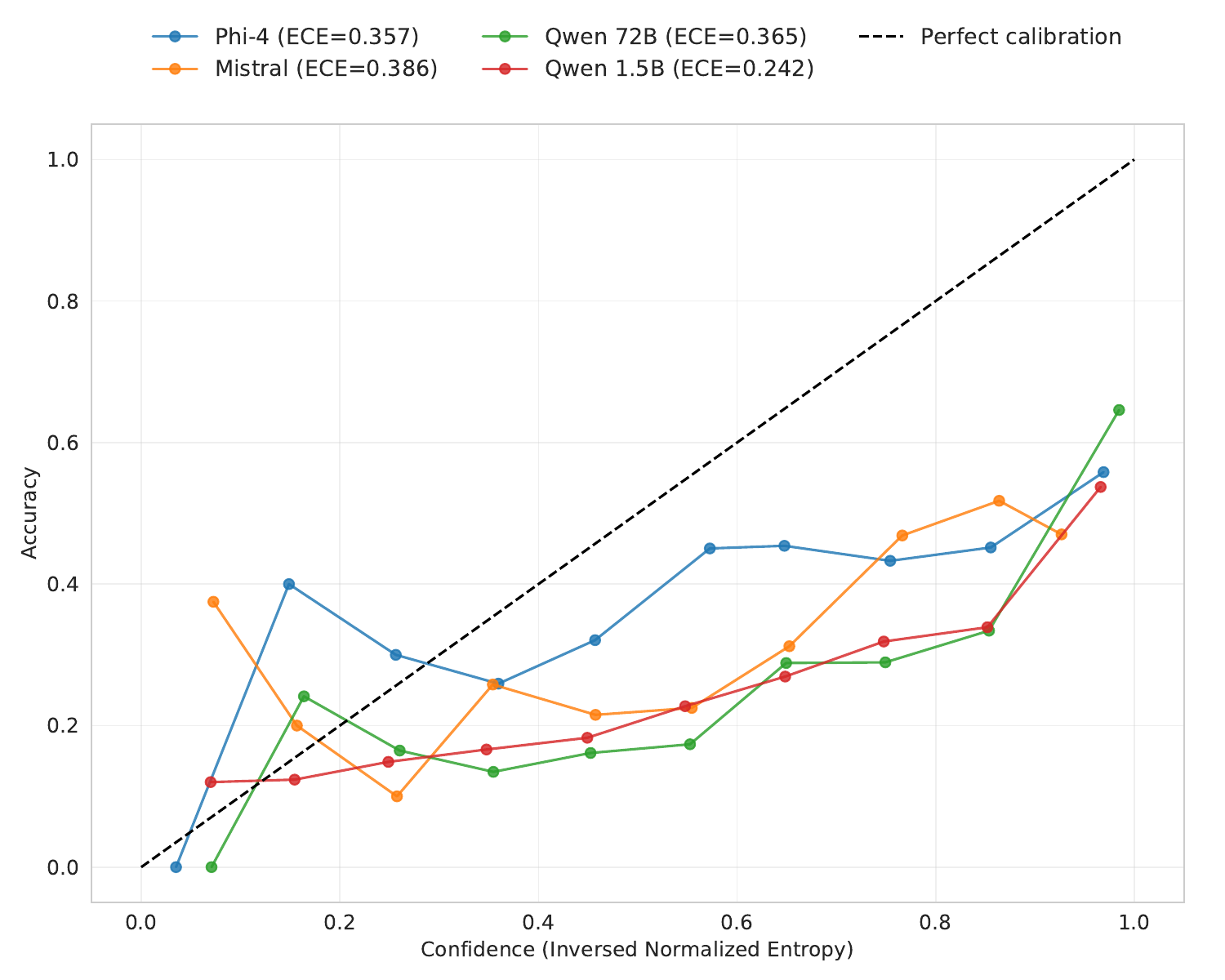}
    \caption{Calibration of LLMs based on their entropy}
    \label{fig:models_compare_calibration}
\end{figure}

\paragraph{Calibration assessment via entropy.}
Figure \ref{fig:models_compare_calibration} illustrates calibration curves for four language models, computed using inverted normalized entropy. The analysis reveals systematic deviations between self-reported certainty and empirical accuracy, with distinct patterns across architectures:

Mistral exhibits unstable calibration dynamics, characterized by erratic accuracy in low-confidence regions. Despite moderate confidence estimates, its accuracy fluctuates unpredictably, even peaking anomalously in the lowest entropy bin. This contrasts with its pronounced underperformance in high-confidence regimes, where empirical accuracy consistently trails reported certainty—a limitation indicative of architectural weaknesses in uncertainty quantification.

Qwen 72B, despite its large scale, demonstrates inconsistent calibration. High-confidence predictions show weak alignment with accuracy, while mid-range entropy bins display substantial confidence-accuracy gaps.

Qwen 1.5B emerges as an outlier, surpassing larger counterparts in calibration quality. Its confidence-accuracy relationship progresses monotonically across entropy bins, with minimal extreme deviations.

These results underscore systematic overconfidence as a pervasive issue, especially in high-certainty regimes where all models significantly overestimate their reliability.

\section{Conclusion and discussion}

Risks in LLMs involve possible errors in answering questions with high confidence.
One possible way to mitigate this problem is by highlighting those questions that, with high probability, would be answered incorrectly.
We suggest using uncertainty estimations for LLMs to identify possible errors.
For a more detailed analysis, the paper examines various subject areas, question types, and specific requirements for knowledge and reasoning.
We explored how different uncertainty estimates for LLMs highlight potential errors of a model and examined various aspects of problems in QA to identify what type of problems a specific uncertainty estimate can highlight.
Our focus was on reliable and popular methods, including a model as a judge and an entropy-based approach.

Experimental evidence suggests that different uncertainty estimates perform differently across diverse subject areas.
Our findings also demonstrate that the key factor is the varying composition of question sets within a subject area collected for a specific subject area: entropy performs much better when no reasoning or little reasoning is required. Thus, it primarily highlights the lack of the required knowledge. 
The conclusion also depends on the model used: for reasoning-oriented, such as Qwen-72B, entropy better highlights the possibility of an error compared to Mistral.

\bibliography{anthology,custom}
\bibliographystyle{amsplain}

\appendix
\section{Prompts} \label{prompts}

This section provides prompts used for Model-as-judge approaches: for numerical estimation of complexity for MASJ in Table~\ref{tab:num_masj_prompt},
for nominal estimation of complexity for MASJ in Table~\ref{tab:nom_masj_prompt},
and for reasoning level estimate in Table~\ref{tab:reasoning_level_prompt}.

\begin{table}[H]
    \small
    \centering
    \caption{Prompt for Numerical model-as-judge}
    \label{tab:num_masj_prompt}
    \begin{tabular}{p{11cm}}
    \hline
    You are an expert in the topic of the question. Please act as an impartial judge and evaluate the complexity of the multiple-choice question with options below. Begin your evaluation by providing a short explanation. Be as objective as possible. After providing your explanation, you must not answer the question. You must rate the question complexity as a number from 0 to 1 following the following scale as a reference: \\
    high\_school\_and\_ easier - 0.0-0.22, \\
    undergraduate\_easy - 0.2-0.4, \\ 
    undergraduate\_hard - 0.4-0.6, \\ 
    graduate - 0.6-0.8, \\ postgraduate - 0.8-1.0. \\
    You must return the complexity by strictly following this format: "  
    [[complexity]]", \\ for example: "Your explanation... Complexity: [[0.55]]", which corresponds to a hard question at the undergraduate level.
    \\
    \hline 
    \end{tabular}
\end{table}

\begin{table}[H]
    \small
    \centering
    \caption{Prompt for Nominal model-as-judge}
    \label{tab:nom_masj_prompt}
    \begin{tabular}{p{11cm}}
    \hline
    You are an expert in the topic of the question. Please act as an impartial judge and evaluate the complexity of the multiple-choice question with options below. Begin your evaluation by providing a short explanation. Be as objective as possible. After providing your explanation, you must not answer the question. You must rate the question complexity by strictly following the scale: \\
    high\_school\_and\_easier, \\ 
    undergraduate\_easy, \\ 
    undergraduate\_hard, \\ 
    graduate, \\ 
    postgraduate. \\
    You must return the complexity by strictly following this format: "[[complexity]]", \\ 
    for example: "Your explanation... Complexity: [[undergraduate]]", which corresponds to the undergraduate level.
    \\
    \hline 
    \end{tabular}
\end{table}

\begin{table}[H]
    \centering
    \small
    \caption{Prompt for Reasoning level estimate}
    \label{tab:reasoning_level_prompt}
    \begin{tabular}{p{11cm}}
    \hline
    You are an expert in the topic of the question. Please act as an impartial judge and evaluate the complexity of the multiple-choice question with options below. \\
    Begin your evaluation by providing a short explanation. Be as objective as possible. After providing your explanation, you must not answer the question. \\
    You must rate the question complexity by strictly following the criteria: \\
    1) [[Requires knowledge]] - do we need highly specific knowledge from the domain to answer this question? Valid answers: yes, no; \\
    2) [[Requires reasoning]] - do we need complex reasoning with multiple logical steps to answer this question? Valid answers: yes, no; \\
    3) [[Number of reasoning steps]] - how many reasoning steps do you need to answer this question? Valid answers: low, medium, high. \\
    Your answer must strictly follow this format:  
    "[[Requires knowledge: answer]] [[Requires reasoning: answer]] [[Number of reasoning steps: answer]]". \\
    Example 1: "Your explanation... [[Requires knowledge: yes]] [[Requires reasoning: no]] [[Number of reasoning steps: low]]". \\
    Example 2: "Your explanation... [[Requires knowledge: no]] [[Requires reasoning: yes]] [[Number of reasoning steps: high]]". \\
    Example 3: "Your explanation... [[Requires knowledge: yes]] [[Requires reasoning: yes]] [[Number of reasoning steps: medium]]". \\
    Example 4: "Your explanation... [[Requires knowledge: no]] [[Requires reasoning: no]] [[Number of reasoning steps: low]]". \\
    \hline
    \end{tabular}
\end{table}

\section{Question examples with answers}
\label{sec:question_examples}

This section provides examples of questions included in the MMLU-Pro dataset on Law, Psychology, and Engineering in Tables~\ref{tab:law_example}, \ref{tab:psychology_example}, and~\ref{tab:engineering_example}, correspondingly. 

\begin{table}[H]
    \small
    \centering
    \caption{Example of a question from Law category}
    \label{tab:law_example}
    \begin{tabular}{p{11cm}}
    \hline
    Which of the following criticisms of Llewellyn's distinction between the grand and formal styles of legal reasoning is the most compelling? \\
    1. There is no distinction between the two forms of legal reasoning. \\
    2. Judges are appointed to interpret the law, not to make it. \\
    3. It is misleading to pigeon-hole judges in this way. \\
    4. Judicial reasoning is always formal. \\
    \hline
    \end{tabular}
\end{table}

\begin{table}[H]
    \small
    \centering
    \caption{Example of a question from Psychology category}
    \label{tab:psychology_example}
    \begin{tabular}{p{11cm}}
    \hline
    A 66-year-old client who is depressed, has rhythmic hand movements, and has a flattened affect is probably suffering from: \\
    1. Huntington's disease \\
    2. Creutzfeldt-Jakob disease \\
    3. Multiple Sclerosis \\
    4. Alzheimer's disease \\
    5. Parkinson's disease \\
    6. Vascular Dementia \\
    7. Frontotemporal Dementia \\
    8. Schizophrenia \\
    9. A right frontal lobe tumor \\
    10. Bipolar Disorder \\
    \hline
    \end{tabular}
\end{table}

\begin{table}[H]
    \small
    \centering
    \caption{Example of a question from Engineering category}
    \label{tab:engineering_example}
    \begin{tabular}{p{11cm}}
    \hline
    A cumulative compound motor has a varying load upon it which requires a variation in armature current from 50 amp to 100 amp. \\
    If the series-field current causes the air-gap flux to change by 3 percent for each 10 amp of armature current, find the ratio of torques developed for the two values of armature current. \\
    1. 2.26 \\
    2. 0.66 \\
    3. 3.95 \\
    4. 1.00 \\
    5. 2.89 \\
    6. 1.75 \\
    7. 4.12 \\
    8. 1.15 \\
    9. 0.87 \\
    10. 3.40 \\
    \hline
    \end{tabular}
\end{table}

\section{Entropy Distributions}
\label{sec:entropy_distributions }

\begin{figure}[H]
\centering
\includegraphics[width=0.95\textwidth]{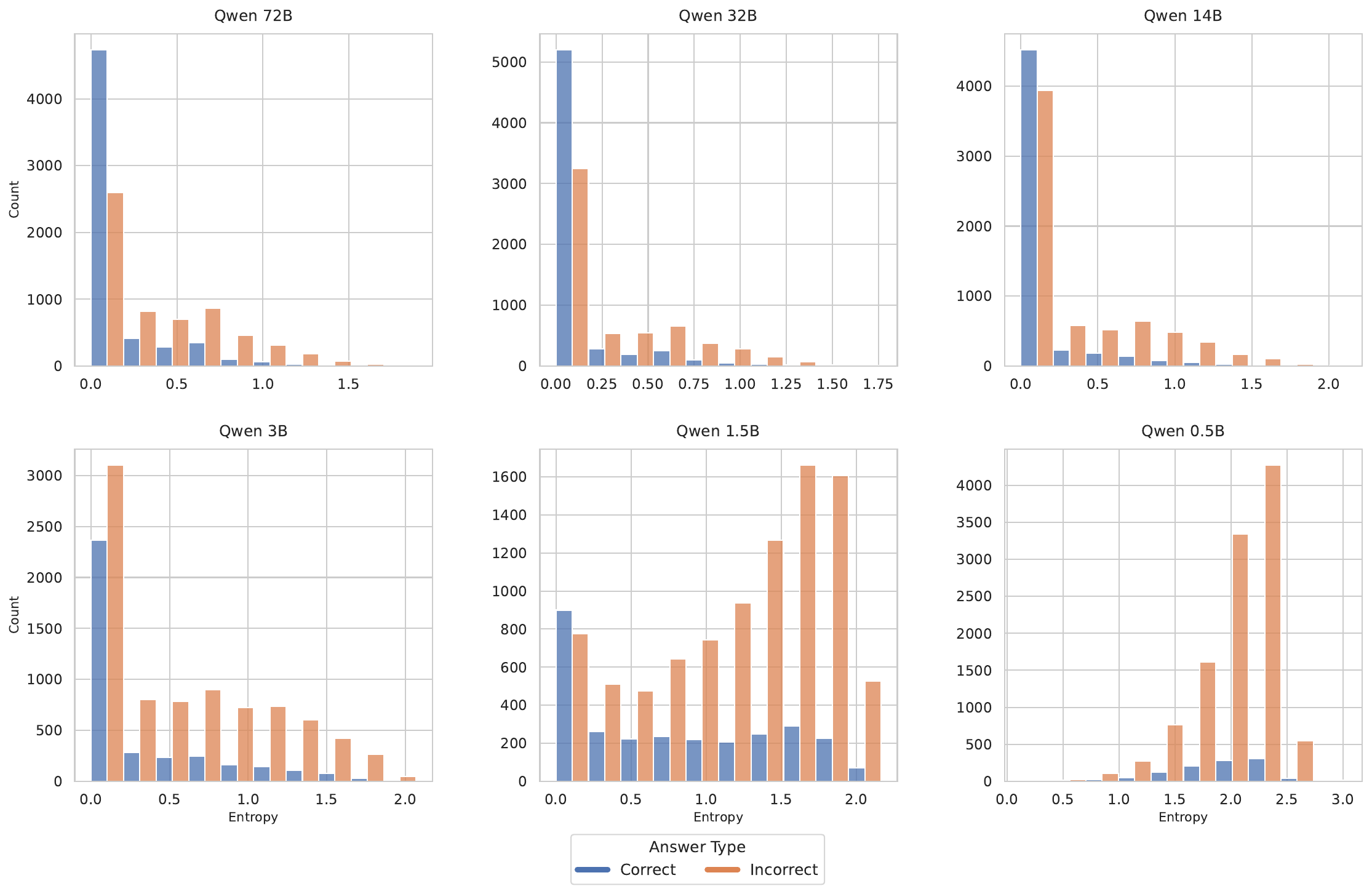}
\caption{Distribution of entropy values for Qwen models (72B, 32B, 14B, 3B, 1.5B, 0.5B), stratified by answer correctness. For larger models (72B, 32B), correct answers (blue) exhibit a pronounced left skew toward low entropy, indicating higher confidence in accurate predictions. Smaller models (1.5B, 0.5B) show flatter distributions, with less separation between correct and incorrect (orange) entropy values. Notably, the 72B variant demonstrates near-zero entropy peaks for correct responses, aligning with findings from Section 4.2.} 
\label{fig:entropy_distributions }
\end{figure}

This section complements the analysis in Section 4.3 by providing full entropy distributions for the Qwen model family. Figure~\ref{fig:entropy_distributions } reveals two key trends:

Model scale correlates with entropy separation: Larger models (72B, 32B) show clearer divergence between correct (low entropy) and incorrect (higher entropy) predictions, while smaller variants (3B, 0.5B) exhibit overlapping distributions.

\end{document}